\documentclass[11pt]{article}

\usepackage[preprint]{acl}

\usepackage{times}
\usepackage{latexsym}
\usepackage[T1]{fontenc}
\usepackage{twemojis}

\usepackage[utf8]{inputenc}
\usepackage{amsmath}

\usepackage{microtype}

\usepackage{inconsolata}

\usepackage{graphicx}
\usepackage{booktabs}
\usepackage[most,skins,theorems]{tcolorbox}
\usepackage{xcolor}
\usepackage[table]{xcolor}
\tcbuselibrary{listings, breakable}
\tcbset{
    graybox/.style={
        colback=gray!20,  
        colframe=black,   
        boxrule=1pt,      
        arc=4mm,          
        width=\textwidth, 
        before=\vspace{10pt}, 
        after=\vspace{10pt},  
        center             
    }
}
\definecolor{bg_gray}{RGB}{245,245,245}
\definecolor{javagreen}{rgb}{0.25,0.5,0.35} 

\title{
\twemoji[height=2.5ex]{cloud}
MIST: Multimodal Interactive Speech-based Tool-calling \\
Conversational Assistants for Smart Homes
}

\author{Maximillian Chen$^1$\thanks{denotes equal contribution. YJ is the corresponding author. MC is now at Google and AP is now at Apple.
}
, 
  Xuanming Zhang$^{1*}$,
  Michael Peng, \\
  \textbf{Zhou Yu}$^{1}$,
  \textbf{Alexandros Papangelis},
  \textbf{Yohan Jo}$^2$
  \\
  $^1$Columbia University, $^2$Seoul National University \\
  \texttt{\{maxchen, billyzhang\}@cs.columbia.edu, yohan.jo@snu.ac.kr} \\
}

\begin{document}
\maketitle
\begin{abstract}
The rise of Internet of Things (IoT) devices in the physical world necessitates voice-based interfaces capable of handling complex user experiences. While modern Large Language Models (LLMs) already demonstrate strong tool-usage capabilities, modeling real-world IoT devices presents a difficult, understudied challenge which combines modeling spatiotemporal constraints with speech inputs, dynamic state tracking, and mixed-initiative interaction patterns. We introduce MIST (the \textbf{M}ultimodal \textbf{I}nteractive \textbf{S}peech-based \textbf{T}ool-calling Dataset), a synthetic multi-turn, voice-driven code generation task that operates over IoT devices. We find that there is a significant gap between open- and closed-weight multimodal LLMs on MIST, and that even frontier closed-weight LLMs have substantial headroom. We release MIST and an extensible data generation framework to build related datasets in order to facilitate research on mixed-initiative voice assistants which reason about physical world constraints.\footnote{ \url{billyzhang24kobe.github.io/mist-smarthome}}
\end{abstract}

\section{Introduction \& Related Work}
\begin{figure}[ht]
    \centering
    \includegraphics[width=0.99\linewidth]{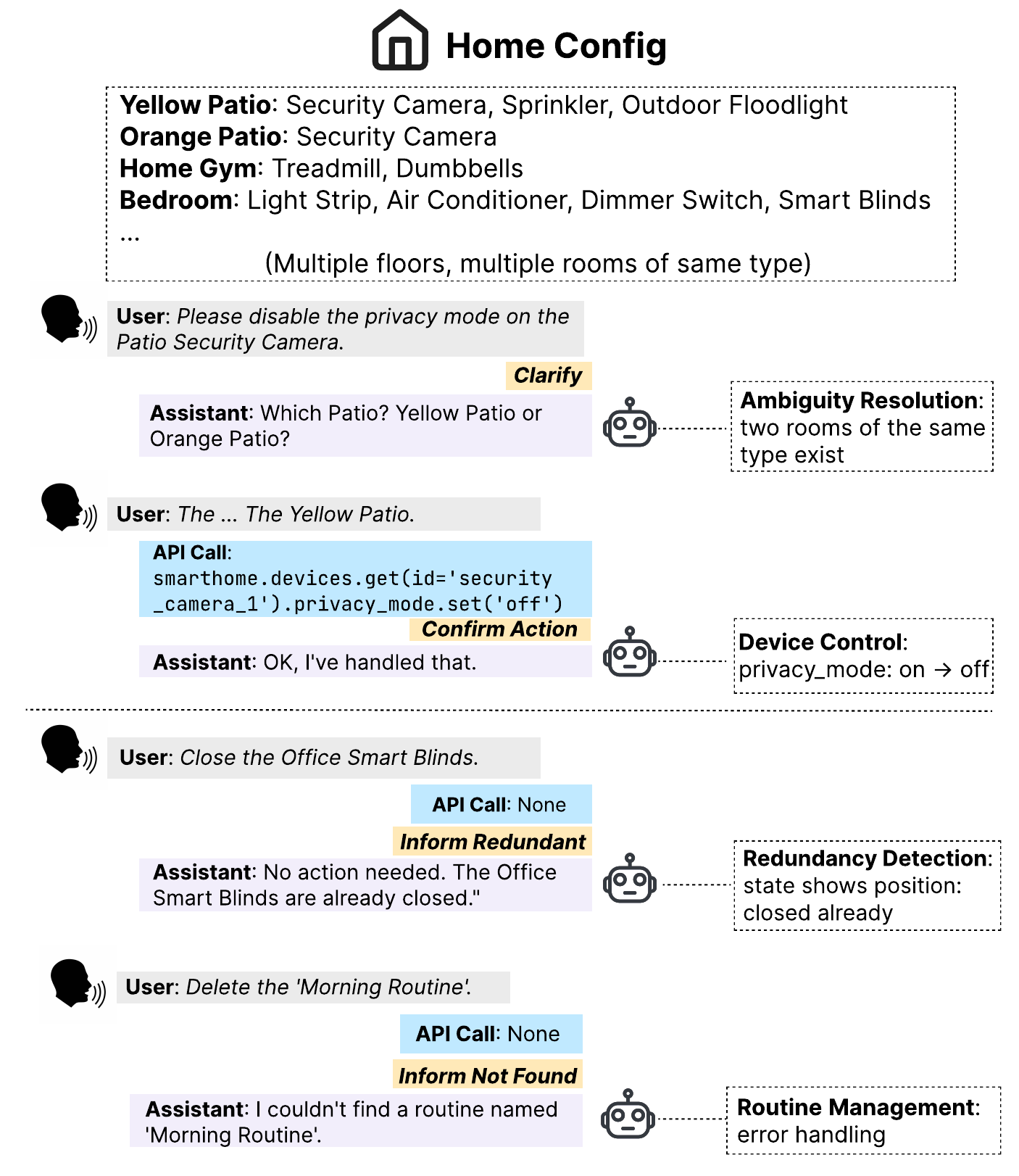}
    \caption{\textbf{Example conversation from MIST}. Users issue voice commands with natural disfluencies and varied accents. The assistant must generate structured API calls while managing ambiguity, corrections, redundancy, and stateful device tracking across turns.}
    \label{fig:teaser}
    \vspace{-6mm}
\end{figure}

The Internet of Things serves as an interface between the physical and the virtual world through a network of interconnected devices. IoT adoption continues to accelerate with recent advances in bringing large language models to virtual assistants (e.g. Alexa+, Gemini for Home), and by 2030 there are expected to be nearly 40 billion connected IoT devices~\cite{iji2024iot}. As these systems include increasingly complex capabilities, rigid rule-based interfaces become insufficient. Multimodal Large Language Models (MLLMs) capable of reasoning over both \textit{spoken} and \textit{textual} modalities offer a promising path toward developing agents that can navigate diverse physical constraints and user interaction patterns. 

\begin{figure*}[ht]
    \centering
    \includegraphics[width=\linewidth]{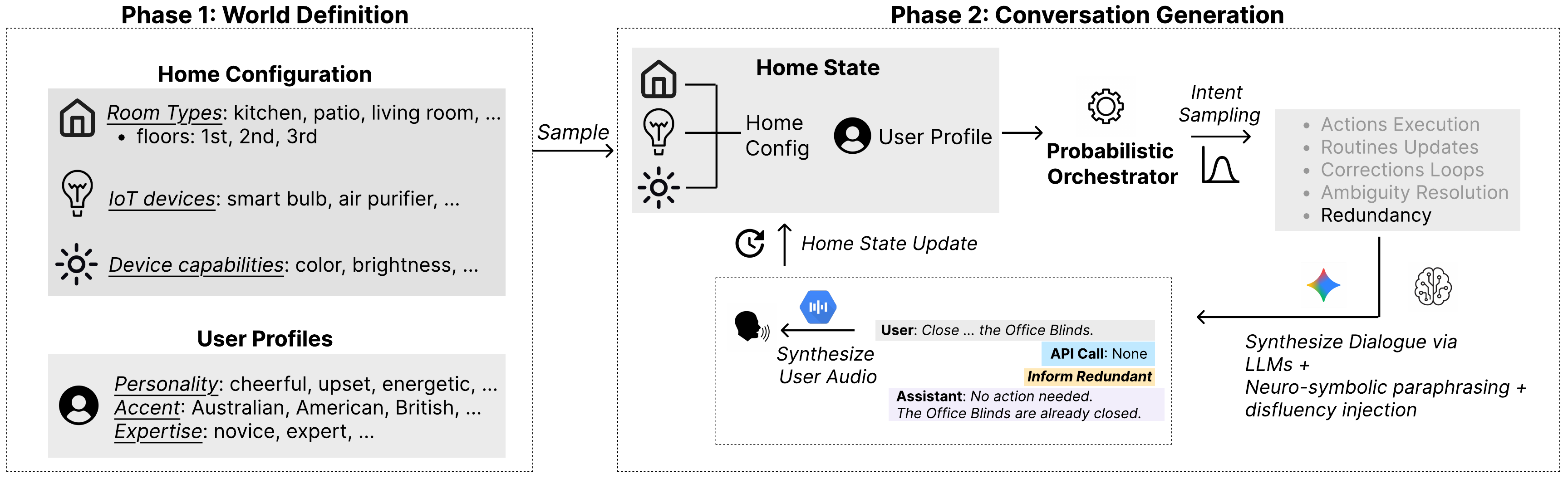}
    \caption{\textbf{Overview of the data generation framework to construct MIST.} We first sample from diverse set of possible user personas, IoT devices, and rooms to form home configurations, then repeatedly sample valid conversational actions and tool calls conditioned on these configurations to form goal-oriented conversations.
    }
    \label{fig:overview}
    \vspace{-4mm}
\end{figure*}

Developing a modern multimodal conversational assistant for real-world IoT devices necessitates going beyond traditional Task-Oriented Dialogue (TOD) tasks such as slot filling and intent detection~\cite{coucke2018snips, hemphill1990atis,schuster2019cross}. Modern challenges include managing a stateful representation of the physical world~\cite{rivkin2024aiot}, executing tool calls~\cite{goel2023presto} to orchestrate actions across various devices, modeling multi-turn conversational histories~\cite{budzianowski2018multiwoz, rastogi2020towards}, and maintaining robustness when presented with disfluent users~\cite{goel2023presto, qin2024toolllm}. In this paper, we build on a rich history of work in TOD and conversational task synthesis~\cite{
bae2022building,qian2025bottom}, alongside growing bodies of work in digital text-based tool-calling~\cite{qin2024toolllm} and speech-based TOD~\cite{zhang2023groundialog,faisal2021sd,si2023spokenwoz}. We introduce \textbf{MIST} (\textbf{M}ultimodal \textbf{I}nteractive \textbf{S}peech-based \textbf{T}ool-calling Dataset), a novel benchmark task requiring MLLMs to jointly model spoken requests in multi-turn dialogues with mixed-initiative conversation dynamics, while understanding API calls with physical world implications and spatiotemporal constraints. To construct MIST, we created a neuro-symbolic data generation framework.

\section{MIST Overview}
MIST features 10,000 conversations with 88.1 hours of spoken dialogue. MIST includes 50 of the most common unique IoT devices spanning 27 unique capabilities/API functions, both sourced from online articles~\cite{bostonautomations2025, ecosmarthomepros2025,bhhs2025}. Each conversation features an average of 5.6 user turns. As in Figure~\ref{fig:teaser}, each conversation involves a user asking a virtual assistant to interact with physical IoT devices.

\subsection{Data Generation Framework}
Figure~\ref{fig:overview} presents an overview of the data generation framework for MIST. For the first phase of the data generation framework, we start by defining a set of possible values for each of these three. We define ``room types'' (e.g., ``kitchen'' or ``patio'') according to an ontology defined in Table~\ref{tab:hierarchy}. Each of these rooms is mapped to a set of plausible IoT devices. Each IoT device has its own unique capabilities (e.g., ``color'' or ``brightness'' on a smart bulb), which can be interacted with using function calls. The supported IoT devices with their capabilities and placement constraints are defined in Table~\ref{tab:device_catalog}. We lastly define possible values for user traits in terms of behaviors personalities (e.g., ``cheerful''; see Table~\ref{tab:behavioral_profile}), expertise (e.g., ``novice''; see Table~\ref{tab:behavioral_profile}), speaking accent (e.g., ``Australian''), speaking pitch, speaking rate, and equipment noise (which maps to Gaussian noise; see Table~\ref{tab:acoustic_profile}).

The second phase entails conversation generation managed by a probabilistic orchestrator. For each conversation, the framework samples a unique home configuration and a consistent user profile. The home configuration parameterizes a \textit{Home State} object which serves as a "Digital Twin" of the physical-world device state~\cite{vanderhorn2021digital} that tracks the real-time status of every device capability and routine. The orchestrator probabilistically samples a target interaction intent at each turn. Once an intent is selected, the system performs a symbolic check against the Home State to ground the interaction. Our framework supports six core interaction patterns (i.e. dialogue actions).

\textbf{1) Action Executions:} Users request an action to be executed over devices in real-time (e.g., ``turn off everything on the \textit{second floor}'') and the agent must identify that it is a valid request and produce the correct tool call. \textbf{2) Routine Updates:} Users may request combinations of actions, triggers, and conditions, which can be created, updated, and deleted (e.g. ``turn on the patio light on weekends at 7am''), and the agent must identify whether it is valid and produce the correct call to update the Smart Home's routine manager. \textbf{3) Correction Loops:} The agent applies a user-requested correction (e.g., ``actually, I meant to set the volume to 30'') through multiple tool calls while ``undoing'' previous actions if necessary. \textbf{4) Ambiguity Resolution:} The orchestrator identifies potential collisions at three levels: device name duplicates, room type ambiguity (e.g., two bedrooms), or intra-room device type duplicates. In these cases, it generates a clarification sub-dialogue where the user poses an underspecified request and the agent must ask a clarifying question\footnote{We assign randomized colors to differentiate rooms of the same type (e.g., ``Blue Bedroom'' vs. ``Red Bedroom'')} (e.g., in Figure~\ref{fig:teaser}, there are multiple rooms of the same type). \textbf{5) Redundancy:} The user may ask for a redundant ``no-op'' request and the agent needs to be capable of recognizing and rejecting them by evaluating the current Home State. \textbf{6) Status Updates:} The user may ask for the current status of the smart home and the agent should form a tool call to retrieve the state of all of the devices. After each of these interactions, the home state is updated based on the code execution. 

Each of these interaction patterns map to a pair containing a fixed user-side dialogue action and an ``optimal'' agent-side dialogue action. Both of which have default templated utterances. The user-side dialogue is paraphrased according to the sampled behavioral traits for that conversation using Gemini 2.5 Flash-Lite. To reflect naturalistic interaction, a rule-based injector then randomly adds speech disfluencies, including word repetitions and revisions~\cite{shriberg1994preliminaries,passali-etal-2022-lard}. Finally, the text is synthesized into audio using the Google Cloud TTS API according to the sampled acoustic profile, with Gaussian noise injected to simulate recording noise (following \citet{chen2025data}). Implementation details are in Appendix~\ref{sec:appendix_value_spaces}.

To vet the dataset quality, we randomly sampled 300 examples and asked expert annotators to listen to the spoken request and read the existing smart home context. The annotators were tasked with verifying correctness with respect to the dataset's stated golden dialogue actions and tool calls. We find that over 92\% of both the dialogue actions and the proposed tool calls are correct, and there is over 90\% agreement between annotators for these tasks. Full human evaluation details are in Appendix~\ref{sec:humaneval}.

\section{Experiments}
In MIST, the following text inputs are provided to an MLLM: the smart home layout (including all IoT devices with their capabilities), the existing Home State, and existing conversation history. The MLLM also receives the user's current request (i.e., the target) as speech. The prompt used to aggregate each of these inputs is in Appendix~\ref{fig:conversational_inputs}.

\paragraph{Evaluation} Models are evaluated along two dimensions. First is \textbf{Code Intelligence}, given in \textit{Execution Match} (percentage of turns where the generated tool calls result in the correct final home state) and \textit{Exact Match} (character-level match of the generated code), as in \citet{yu2019cosql}. These metrics are computed for examples that require tool calls. The second is \textbf{Conversational Intelligence}: the agent's ability to recognize ambiguities, redundancies, and other phenomena by producing responses with the correct \textit{dialogue action}. We measure the Macro F1 and Accuracy of the inferred actions (implementation details in Appendix \ref{sec:appendix_eval_details}).
This reflects the Action-level evaluation setting proposed in ~\citet{chen2025learning} and is measured using Macro F1 and Accuracy.

\paragraph{Baselines} We contextualize MLLM performance using several baselines. For code generation, we use a baseline where we use the initial home state and compute the ``execution match'' using this state for every turn of the conversation (``Initial State''). We also consider a baseline which assumes no change from the previous turn's home state (``Previous State''). For conversational intelligence, we present a baseline that assumes that the candidate response always follows the most common dialogue action in MIST (``Constant Prediction'').

\paragraph{Models} We consider several competitive open-weight MLLMs: Qwen Audio~\cite{chu2023qwen}, Qwen 2 Audio~\cite{chu2024qwen2}, Soundwave~\cite{zhang2025soundwave}, and Qwen 3 Omni~\cite{xu2025qwen3}. We also evaluated a frontier closed-weight model family: Gemini 2.5 Flash-Lite, Flash, and Pro~\cite{comanici2025gemini}.

\begin{figure*}[h]
    \centering
    \includegraphics[width=0.49\linewidth]{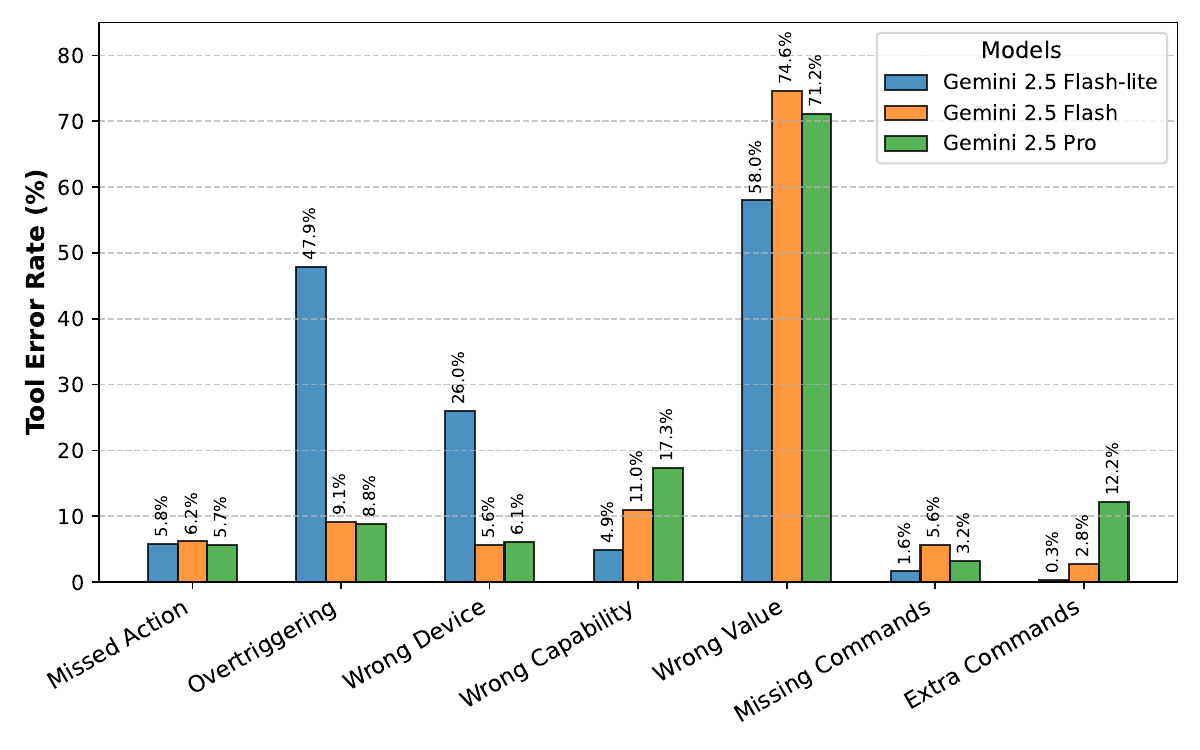}
    \includegraphics[width=0.49\linewidth]{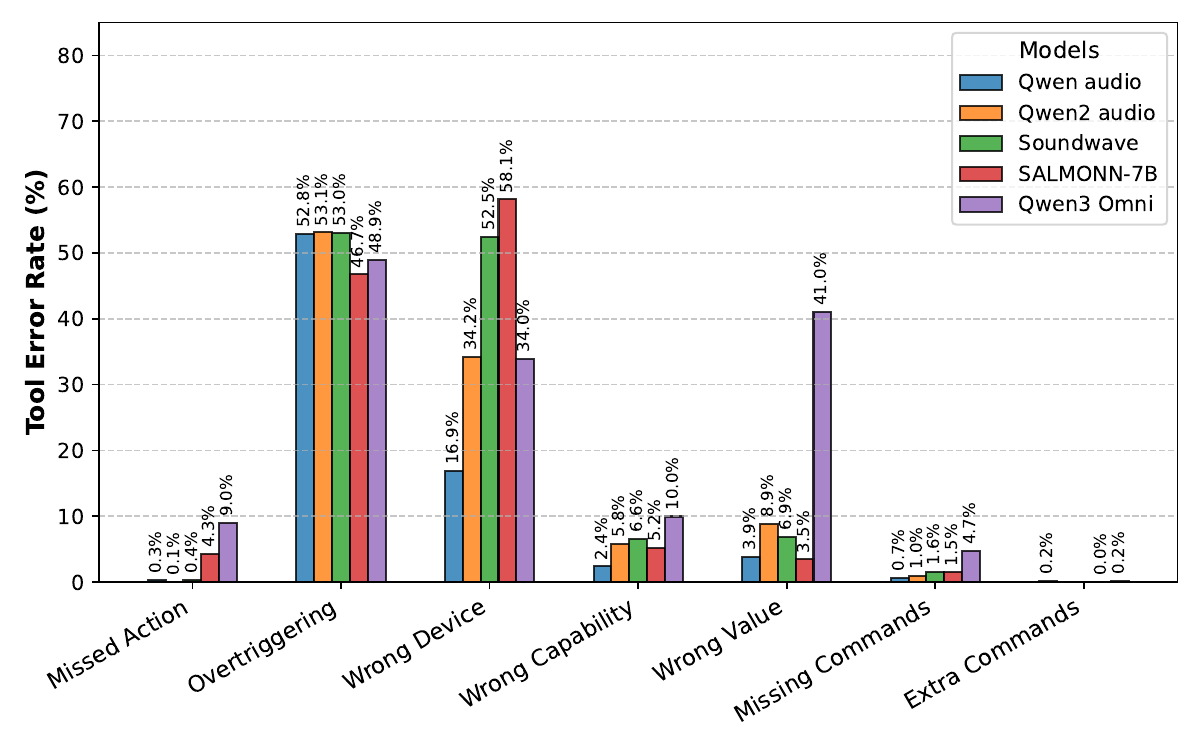}
    \caption{\textbf{Error analysis characterizing the types of errors by proportion for each MLLM.} 
    The most common tool execution error for frontier models is selecting the `Wrong Value`, whereas open-weight models struggle triggering a tool call at the wrong time or targeting the wrong device. 
    }
    \label{fig:error_analysis}
    \vspace{-4mm}
\end{figure*}
\subsection{Results \& Discussion}
\label{sec:main_results}

\begin{table}[h]
\centering
\small
\begin{tabular}{lcc}
\toprule
\textbf{Model} & \textbf{Exec Match} & \textbf{Exact Match} \\
\midrule
Initial State & 35.33 & --- \\
Previous State & 71.62 & --- \\
\midrule
Qwen Audio & 57.19 & 1.18 \\
Qwen 2 Audio & 60.94 & 0.45 \\
SALMONN 7B & 48.76 & 1.01 \\
Soundwave & 49.83 & 2.26\\
Qwen 3 Omni & 59.86 & 47.19 \\
\midrule
Gemini 2.5 Flash-Lite & 57.30 & 42.50 \\
Gemini 2.5 Flash & 78.61 & 63.95 \\
Gemini 2.5 Pro & \textbf{79.53} & \textbf{65.56} \\
\bottomrule
\end{tabular}
\caption{\textbf{Code Generation} results indicate Gemini 2.5 Pro achieves the strongest Exact Match, with a substantial gap over leading open-weight models.}
\label{tab:code_results}
\vspace{-4mm}
\end{table}
\paragraph{Code Intelligence}
Table~\ref{tab:code_results} shows there are clear gaps between closed-weight frontier MLLMs and leading open-weight audio models. Open-weight models achieve moderate Execution Match scores (ranging from 48.76\% to 60.94\%), yet all but Qwen 3 Omni fail almost entirely on the Exact Match metric ($\le 2.26\%$). The ``Previous State'' baseline reveals that in 71.6\% of examples, performing an action over the IoT devices is not required (e.g., the agent should elicit more information or reject the request). The Code Intelligence error analysis in Figure~\ref{fig:error_analysis} shows at least 46\% of the erroneous function calls involve ``overtriggering'' for all open-weight MLLMs, meaning the agent performs an unnecessary code action. The second most common error for open-weight MLLMs is targeting the ``wrong device.'' This suggests models are not effective at understanding complex contexts which may feature similar devices, which has \textit{serious physical world implications} (e.g. leaving the wrong door unlocked, turning on the wrong oven).  
In contrast, the closed-weight MLLMs achieve decent performance. Gemini 2.5 Pro achieves the strongest performance with a 79.53\% Execution Match and a 65.56\% Exact Match. The overall number of errors for closed-weight models is much lower, evidenced by the lower rates of overtriggering and selecting the wrong device. Instead, the most common error type is producing the ``wrong value'' in code (e.g., setting the speaker to the incorrect volume setting). Lastly, we also see model performance seems to improve with model scale, suggesting that the task is climbable and there is substantial opportunity to bridge the cross-modal reasoning capabilities between open- and closed-weight MLLMs.

\begin{table}[t]
\centering
\small
\begin{tabular}{lcc}
\toprule
\textbf{Model} & \textbf{Macro F1} & \textbf{Accuracy} \\
\midrule
Constant Prediction & \phantom{0}9.13 & 37.74\\
\midrule
Qwen Audio & \phantom{0}9.14 & 18.92 \\
Qwen 2 Audio & \phantom{0}6.21 & 22.35 \\
SALMONN 7B & \phantom{0}6.44 & 21.23 \\
Soundwave & \phantom{0}6.80 & 13.38 \\
Qwen 3 Omni & 14.54 & 28.64 \\
\midrule
Gemini 2.5 Flash-Lite & 15.37 & 31.55 \\
Gemini 2.5 Flash & 42.17 & 63.57 \\
Gemini 2.5 Pro & \textbf{46.00} & \textbf{66.73} \\
\bottomrule
\end{tabular}
\caption{\textbf{Conversational Intelligence} based on inferred dialogue actions in terms of F1 and Accuracy.}
\label{tab:intent_results}
\vspace{-6mm}
\end{table}

\paragraph{Conversational Intelligence}
Table~\ref{tab:intent_results} demonstrates models' mixed-initiative interaction skills by assessing whether they are able to correctly identify when to confirm an action request, elicit clarifying information from a user, and more. We see that open-weight models struggle severely with producing the right conversational action, posting F1 scores which underperform a constant prediction baseline (9.13 F1). This suggests that current open-weight MLLMs cannot reliably interpret the smart home context to determine when to ask for clarification or reject a redundant request. The Gemini 2.5 models perform substantially better, with Pro achieving 46.00 F1 and 66.73\% Accuracy. However, Figure~\ref{fig:dialogue_error_analysis} shows that Gemini 2.5 Pro still fails to recognize 73.0\% of cases where the golden action is to confirm a valid request. The large headroom even among frontier models underscores the inherent difficulty of the MIST benchmark. 

\section{Conclusion}
MIST is a novel benchmark for MLLMs' ability to act as code-generating agents which interpret complex user intents with spatial constraints. We find that MIST is a valuable metric to climb on, given the large gap between the abilities of open-weight and closed-weight models and the remaining headroom for frontier closed-weight models. Coupled with its extensible data generation framework which can be used to produce synthetic training data, MIST will serve as a resource to accelerate the development of open-source MLLMs and agentic experiences for the physical world.

\bibliography{custom}
\clearpage
\appendix
\setcounter{figure}{0}
\setcounter{table}{0}
\setcounter{equation}{0}
\renewcommand{\thetable}{A\arabic{table}}
\renewcommand{\thefigure}{A\arabic{figure}}
\renewcommand{\theequation}{A\arabic{equation}}

\section{Additional Related Work}
Many recent efforts have proposed related efforts to assess MLLM speech understanding. \citet{yang2026saynext} introduce SayNext-Bench to assess the ability of multimodal large language models to accurately predict a speaker's next conversational utterance by leveraging non-verbal contextual cues. Other works focus on a similar setting called spoken question answering. \citet{chen2024voicebench} proposes VoiceBench to evaluate the general knowledge, instruction-following capabilities, and safety compliance of LLM-based voice assistants under diverse acoustic and speaker variations. \citet{chen2025data} proposes ASK-QA, a synthetic dataset to understand mixed-initiative speech in conversational QA. Addressing disfluency and naturalness, \citet{liu2025vocalbench} present VocalBench to comprehensively assess speech interaction models across semantic precision, acoustic quality, free-form dialogue, and environmental robustness. To benchmark open-ended speech interactions, \citet{zhang2025wildspeech} develop WildSpeech-Bench, evaluating the end-to-end capabilities of audio LLMs using real-world spoken queries that incorporate speech-specific phenomena.  \citet{yang2025speechr} looks at cross-modal understanding and introduces SpeechR to measure the complex reasoning capabilities of large audio-language models across factual retrieval, procedural inference, and normative judgment.

\section{Additional Experimental Results}
\label{sec:fewshot}

\begin{table}[h]
\centering
\small
\begin{tabular}{lcc}
\toprule
\textbf{Model} & \textbf{Exec Match} & \textbf{Exact Match} \\
\midrule
Qwen Audio & 59.63 & 1.48 \\
Qwen 2 Audio & 60.53 & 0.01 \\
SALMONN 7B & 58.13 & 0.35 \\
Soundwave & 41.60 & 4.22 \\
Qwen 3 Omni & 60.87 & 52.36 \\
\bottomrule
\end{tabular}
\caption{Few-shot code generation results on MIST.}
\label{tab:app_code_results}
\end{table}
We investigated whether the gaps in performance between open-weight and closed-weight models was simply a matter of insufficient domain adaptation. Specifically, at inference time, we experimented with providing randomly sampled 3 few-shot exemplars to each MLLM, ensuring that the evaluation example is never one of the randomly sampled exemplars. The results are presented in Tables \ref{tab:app_code_results} and \ref{tab:app_intent_results}. 

\begin{table}[h]
\centering
\small
\begin{tabular}{lcc}
\toprule
\textbf{Model} & \textbf{Macro F1} & \textbf{Accuracy} \\
\midrule
Qwen Audio & \phantom{0}13.53 & 21.50 \\
Qwen 2 Audio & \phantom{0}4.33 & 25.17 \\
SALMONN 7B & \phantom{0}4.48 & 25.58 \\
Soundwave & \phantom{0}6.22 & 21.13 \\
Qwen 3 Omni & 15.51 & 30.43 \\
\bottomrule
\end{tabular}
\caption{Few-shot intent recognition results on MIST.}
\label{tab:app_intent_results}
\end{table}
The results indicate that few-shot prompting does yield some performance improvements improvements, but is insufficient to bridge the gap between open-weight models and the Gemini 2.5 family remains substantial. Table~\ref{tab:app_code_results} shows that the best-performing open-weight model, Qwen 3 Omni, achieves 52.36\% exact match for code generation, still falling short of Gemini 2.5 Pro's 65.56\% under a zero-shot setting. Table~\ref{tab:app_intent_results} similarly shows that in terms of recognizing the optimal dialogue action, the Qwen 3 Omni achieves the best performance (30.43\% accuracy) whereas Gemini 2.5 Pro achieves 66.73\% zero-shot accuracy. These results underscore both the difficulty of MIST and the persistent performance disparity between open-source and proprietary models.

\section{Data Generation Implementation Details}
\label{sec:appendix_value_spaces}

To construct the MIST dataset, we programmatically sample from predefined value spaces across different stages of the generation pipeline. This structured sampling ensures a highly diverse, expansive, and realistic set of home configurations and user personas. The set of actions that form the basis of the conversation are controlled by configurable probabilities, which determine whether to incorporate redundancies or how handle naming collisions (either by introducing user-side ambiguity or specificity), for instance. 

\begin{table}[h]
\centering
\small
\begin{tabular}{ll}
\toprule
\textbf{Container Hierarchy} & \textbf{Types/Attributes} \\
\midrule
House & --- \\
Floors & 1st, 2nd, 3rd \\
Rooms & Living Room, Bedroom, Kitchen, \\
& Office, Bathroom, Garage, \\
& Dining Room, Home Gym, \\
& Backyard, Home Theater \\
\bottomrule
\end{tabular}
\caption{The hierarchical environment structure used to generate unique world configurations.}
\label{tab:hierarchy}
\end{table}

\begin{table*}[]
\centering
\scriptsize
\begin{tabular}{p{2cm}p{4cm}p{8cm}}
\toprule
\textbf{Category} & \textbf{Device Types} & \textbf{Key Capabilities (Constraints)} \\
\midrule
Lighting & Smart Bulb, Light Strip, Dimmer Switch, Outdoor Floodlight & Power, Brightness, Color, Scene (Floodlight: Outdoor/Garage only) \\ \midrule
Climate & Thermostat, Air Purifier, Ceiling Fan, Smart Blinds, AC & Temp, Mode, Fan Speed, Position (AC: Living Room/Bedroom only) \\ \midrule
Kitchen & Fridge, Oven, Microwave, Coffee Maker, Dishwasher & Mode, Temp, Duration, Brew Strength, Cycle (Kitchen only) \\ \midrule
Entertainment & TV, Soundbar, Speaker, Projector, AV Receiver & Power, Volume, Source, EQ Mode (Projector: Theater only) \\ \midrule
Security & Lock, Camera, Doorbell, Smoke Detector, Garage Opener, Sensors & Lock, Privacy, Chime, Check Status (Doorbell: Living Room only) \\ \midrule
General & Smart Plug, Robot Vacuum, Sprinkler, Pet Feeder, Curtains & Power, Dock, Clean, Dispense, Open/Close (Sprinkler: Outdoor only) \\ \midrule
Home Gym & Treadmill, Smart Scale, Dumbbells & Speed, Incline, Weight, Get Reading (Gym/Bath only) \\ \midrule
Misc & Diffuser, Plant Pot, Smart Mirror & Intensity, Check Moisture, Show Weather \\
\bottomrule
\end{tabular}
\caption{Complete catalog of 50 device types used in MIST, grouped by category. Capabilities define the action space, while constraints limit valid room placements.}
\label{tab:device_catalog}
\end{table*}
\subsection{Environment Simulation}
\label{sec:home_details}
The physical environment of each simulated home is constructed hierarchically. As detailed in Table~\ref{tab:hierarchy}, each environment begins with a root \textit{House} that contains up to three \textit{Floors}, which are further divided into specific \textit{Rooms}. The total number of floors and the room types are each randomly sampled.
Once the container hierarchy is established, rooms are populated with smart devices randomly sampled from the catalog presented in Table~\ref{tab:device_catalog}. This catalog defines 50 distinct device types categorized by their function (e.g., Lighting, Climate, Security). Crucially, device placement is heavily constrained by logical room assignments to maintain realism (e.g., ovens only appear in kitchens, and sprinklers are restricted to outdoor areas).

\begin{table}[h]
\centering
\small
\begin{tabular}{ll}
\toprule
\textbf{Parameter} & \textbf{Sampling Range} \\
\midrule
TTS Accent & \{en-US, en-GB, en-AU\} \\
Pitch Shift & $[-4.0, +4.0]$ semitones \\
Speaking Rate & $[0.85, 1.1] \times$ speed \\
Noise Level & $[0.01, 0.08]$ (Gaussian variance) \\
\bottomrule
\end{tabular}
\caption{Acoustic Profile ($A$) parameters used for TTS synthesis and noise injection.}
\label{tab:acoustic_profile}
\end{table}

\subsection{User Persona and Acoustic Profiles}
\label{sec:profile_details}

\begin{table}[h]
\centering
\small
\begin{tabular}{p{0.95\linewidth}}
\toprule
\textbf{LLM Paraphraser System Prompt} \\
\midrule
You are a persona engine designed to create realistic user dialogue. Your task is to paraphrase a simple, direct command into a more natural and expressive utterance that reflects a specific user profile. The user profile consists of an `expertise level' and a `personality trait'. You must combine these two aspects to create a believable character. For example, a `novice' and `friendly' user might say, `Hi there, could you please do me a favor and turn on the living room light? Thanks so much!' An `expert' and `direct' user might say, `Living room light on.' Only output the final paraphrased command. Do not add any extra conversational text or labels. Ensure the output sounds like natural speech and does not contain code artifacts like underscores or raw boolean values (e.g., use `turn on' instead of `power=true'). \\
\bottomrule
\end{tabular}
\caption{The system prompt used for the Gemini 2.5 Flash-Lite paraphraser to generate persona-driven user utterances based on their assigned behavioral profile.}
\label{tab:paraphraser_prompt}
\end{table}
To simulate a diverse user base, each conversation is grounded in a unique user profile composed of both acoustic and behavioral traits. 
Table~\ref{tab:acoustic_profile} details the acoustic profile parameters used for Text-to-Speech synthesis with the Google Cloud TTS API. By randomizing the TTS accent, pitch shift, speaking rate, and overlaying Gaussian noise, we simulate the varying acoustic challenges and environments that a real-world multimodal assistant would encounter.
Finally, Table~\ref{tab:behavioral_profile} outlines the behavioral profile, which combines one of three expertise levels with a personality trait sampled from over 100 descriptors. Both are randomly sampled. The LLM paraphraser (Gemini 2.5 Flash-Lite) is conditioned on these profiles to produce introduce variance into the user's dialogue, ensuring that MIST captures a broad set of behaviors. The prompt is provided in Table~\ref{tab:paraphraser_prompt}.

\begin{table*}[h]
\centering
\small
\begin{tabular}{l p{0.75\textwidth}}
\toprule
\textbf{Attribute} & \textbf{Value Space} \\
\midrule
Expertise Level & \{Novice, Intermediate, Expert\} \\
Personality Trait & friendly and polite, direct and terse, inquisitive and verbose, hesitant and cautious, cheerful and bubbly, sarcastic and dry, formal and professional, casual and laid back, energetic and enthusiastic, calm and serene, anxious and worried, confident and assertive, creative and imaginative, analytical and logical, empathetic and caring, cynical and skeptical, optimistic and hopeful, pessimistic and gloomy, playful and humorous, serious and somber, adventurous and bold, shy and reserved, grumpy and irritable, patient and understanding, impatient and restless, meticulous and detail oriented, spontaneous and impulsive, nurturing and supportive, independent and self reliant, dramatic and expressive, stoic and unemotional, curious and questioning, whimsical and fanciful, pragmatic and down to earth, artistic and expressive, methodical and organized, chaotic and disorganized, charming and charismatic, aloof and distant, warm and welcoming, cold and indifferent, witty and clever, naive and innocent, worldly and sophisticated, humble and modest, arrogant and boastful, generous and giving, stingy and selfish, honest and transparent, deceptive and secretive, brave and courageous, timid and fearful, loyal and devoted, fickle and unreliable, respectful and deferential, disrespectful and defiant, ambitious and driven, lazy and unmotivated, compassionate and kind, cruel and malicious, diplomatic and tactful, blunt and tactless, forgiving and merciful, vengeful and resentful, flexible and adaptable, rigid and stubborn, grateful and appreciative, entitled and demanding, industrious and hardworking, joyful and merry, melancholy and sad, mysterious and enigmatic, open and straightforward, pensive and thoughtful, quirky and eccentric, rational and level headed, sentimental and nostalgic, tough and resilient, vulnerable and sensitive, zealous and passionate, relaxed and easygoing, intense and focused, goofy and silly, elegant and graceful, awkward and clumsy, philosophical and deep, superficial and shallow, nervous and jittery, bold and daring, apprehensive and hesitant, enthusiastic and eager, apathetic and uninterested, systematic and orderly, impulsive and unpredictable, gentle and tender, harsh and severe \\
\bottomrule
\end{tabular}
\caption{Behavioral Profile attributes used to condition the LLM paraphraser.}
\label{tab:behavioral_profile}
\end{table*}

\begin{table}[h!]
\centering
\small
\begin{tabular}{p{0.95\linewidth}}
\toprule
\textbf{LLM Intent Classifier System Prompt} \\
\midrule
Analyze the following response from a smart home assistant and classify its intent.
        
        Possible Intents:\\
        - affirm: The assistant successfully performed a device action or routine creation/update/deletion.\\
        - clarify: The assistant is asking a clarifying question to resolve ambiguity.\\
        - inform\_redundant: The assistant refused an action because the device/routine is already in that state.\\
        - inform\_not\_found: The assistant could not find a specified routine to delete/update.\\
        - inform\_status: The assistant is reporting the current state of devices.\\
        - apologize\_correct: The assistant is undoing a previous action and performing a new one (correction).
        
        Input:\\
        Assistant Response: "{response\_text}"\\
        Tool Code Generated: "{tool\_code}"
        
        Instructions:\\
        Output ONLY the applicable intent labels from the list above, separated by commas. Do not output anything else. \\
\bottomrule
\end{tabular}
\caption{The system prompt used for the Gemini 2.5 Flash-Lite paraphraser to generate persona-driven user utterances based on their assigned behavioral profile.}
\label{tab:classifier_prompt}
\end{table}

\section{Evaluation Implementation Details}
\label{sec:appendix_eval_details}

We developed an automated evaluation engine that assesses both the correctness of the generated API calls and the semantic correctness of the dialogue actions.

\subsection{Code Generation Evaluation}
Code generation is evaluated along two primary dimensions: exact match and execution  match. This follows prior work such as CoSQL~\cite{yu2019cosql} and AmbigSQL~\cite{chen2025learning}.

\paragraph{Exact Match Accuracy} 
This metric measures strict syntactical adherence. A prediction is marked as an exact match only if the generated \texttt{tool\_code} string is identical to the ground truth code on the character level. This metric rigorously penalizes hallucinated parameters, incorrect device IDs, and missing API calls.

\paragraph{Execution Match Accuracy}
Because multiple valid API sequences can theoretically result in the same physical state, we measure functional correctness by executing the generated code against a local simulator. 
For each conversation turn, the evaluation engine initializes a Home State instance using the smart home configuration and the state snapshot from the preceding turn. The engine parses the model's generated \texttt{tool\_code} and executes the corresponding \texttt{smarthome.devices} and \texttt{smarthome.routines} commands to mutate the simulator's state. 
A turn is considered an Execution Match if, after all predicted code is executed, both the device state dictionary and the routines dictionary of the simulator perfectly match the ground truth \texttt{state\_after\_turn} and \texttt{routines\_after\_turn} provided in the dataset.

\subsection{Dialogue Action Evaluation}
Evaluating the dialogue response with traditional metrics such as BLEU/ROUGE is insufficient for measuring agentic behavior, because they do not capture the nuances of optimal conversational actions (e.g. two sentences can have high token overlap yet differing semantic meaning). Moreover, one can express semantically equivalent phrases using different words. Thus, we evaluate a model's conversational intelligence by formulating it as a multi-label intent classification task using an LLM-as-a-judge following \citet{chen2025learning}.

\paragraph{Intent Classifier Setup}
We utilize \textbf{Gemini 2.5 Flash-Lite} via Vertex AI as a zero-shot intent classifier. To ensure reproducible and deterministic evaluations, the model is configured with a decoding temperature of $T=0.0$ and strict safety filter overrides. 
For each turn, the classifier is provided with the assistant's generated natural language response and the predicted \texttt{tool\_code}. It is prompted to map the response to a subset of six valid dialogue actions: \texttt{confirm\_action}, \texttt{clarify}, \texttt{inform\_redundant}, \texttt{inform\_not\_found}, \texttt{inform\_status}, and \texttt{apologize\_correct}.

\paragraph{Metrics}
The model's predicted intents are compared against the ground truth dialogue actions using set operations. We report three key metrics:
\begin{itemize}
    \item \textbf{Accuracy:} The percentage of turns where the predicted set of intents perfectly equals the ground truth set of intents.
    \item \textbf{Micro F1:} Calculated globally by aggregating the True Positives, False Positives, and False Negatives across all intent predictions across all turns. This provides an overall measure of dialogue action reliability heavily weighted by the most frequent intents.
    \item \textbf{Macro F1:} Calculated by computing the F1 score independently for each of the six intent classes and averaging the results. This ensures that performance on rare but critical mixed-initiative dynamics (e.g., \texttt{inform\_not\_found}, \texttt{apologize\_correct}) is equally weighted.
\end{itemize}

\begin{figure}[h!]
    \centering
        \includegraphics[width=0.99\linewidth]{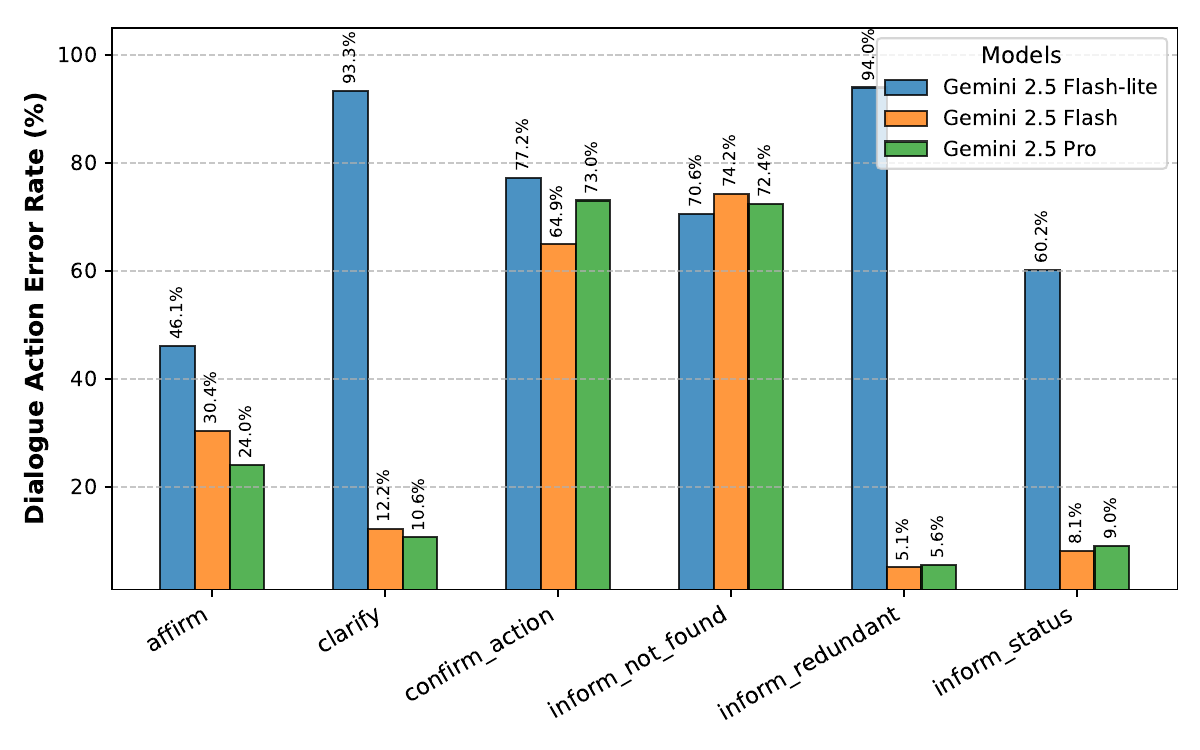}
    \caption{Error analysis of attempted dialogue actions for the Gemini 2.5 model family.}
    \label{fig:dialogue_error_analysis}
\end{figure}
\section{Dialogue Action Error Analysis}
In addition to the analysis of code generation errors in Section~\ref{sec:main_results}, we examine existing gaps in conversational intelligence.

Figure~\ref{fig:dialogue_error_analysis} reveals distinct behavioral profiles across the Gemini model family. The Flash-Lite model acts as an eager agent which prioritizes answering over other conversational strategies. It fails to recognize ambiguity, missing 93.3\% of ambiguous requests and fails to recognize redundancy, missing 94.0\% of redundant requests. Conversely, Gemini 2.5 Flash and Pro are either more capable of recognizing ambiguitiy/redundancy, or more capable of obeying the instruction that the proper behavior is to ask a clarifying question or reject the user's request as opposed to blindly executing an action. We see that Gemini 2.5 Pro only misses 10.6\% of the examples where asking a clarification question was the golden action. However, all models struggle significantly with \texttt{inform\_not\_found} (>70\% error), which suggests that even frontier models have difficulty recognizing when a requested device is entirely absent from the ontology, often attempting to force an execution rather than gracefully rejecting a request.

As for the open-weight models, as indicated in the main results in Table~\ref{tab:intent_results}, the vast majority of attempted dialogue actions are incorrect for all open-weight models.


\section{Human Evaluation}
\label{sec:humaneval}
\begin{figure*}[h!]
    \centering
    \includegraphics[width=0.95\linewidth]{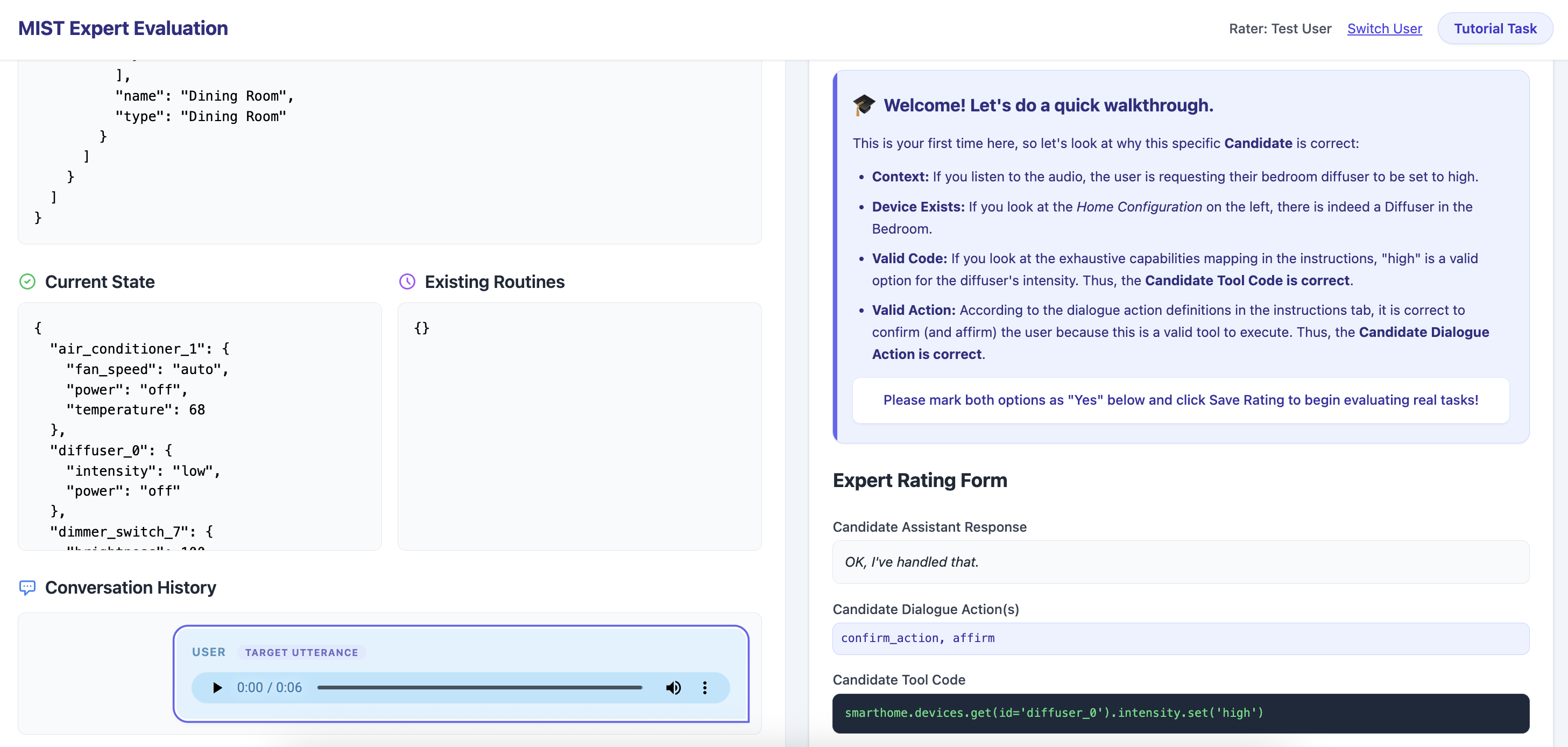}
    \caption{Screenshot of the interface shown to expert human annotators.}
    \label{fig:humaneval}
\end{figure*}
    
\paragraph{Setup Details} To evaluate the quality of MIST data, we conducted an expert human evaluation from a pool of 14 human raters who all have working proficiency in English and at least a graduate background in Computer Science. We randomly sampled 300 examples and randomly assigned three unique raters per example. As shown in Figure~\ref{fig:humaneval}, we ask the raters to determine the correctness of the dialogue action and the code. At the top of the tool, the annotators are also presented with the following instruction: 

\begin{quote}
\small
\textbf{Evaluator Goal:}
Your goal is to judge the correctness of the Candidate Dialogue Action and the Candidate Tool Code.

Please cross-check the Home Configuration, Current State, and Existing Routines below in order to determine the correctness of the action and the code based on the AI's instructions.

\textbf{Assistant Instructions:} See \textit{``MIST System Prompt''} in Section~\ref{sec:prompt_details}.

\textbf{Dialogue Action Definitions:}
When evaluating the candidate action, ensure the generated dialogue action correctly matches the following intent categories:
\begin{itemize}
    \item \texttt{confirm\_action}: Assistant successfully executed a device action or routine change (creation, update, or deletion).
    \item \texttt{clarify}: Assistant asked a question to resolve ambiguity or request missing parameters.
    \item \texttt{inform\_redundant}: Assistant took no action because the target device or routine was already in the requested state (a No-op).
    \item \texttt{inform\_not\_found}: Assistant informed the user that a requested device or routine does not exist.
    \item \texttt{inform\_status}: Assistant answered a question about the current state of the home or specific devices.
    \item \texttt{apologize\_correct}: Assistant reverted a previous mistake and executed the corrected action in response to a user's self-correction.
    \item \texttt{other}: General chitchat or any intent not covered by the primary actions.
\end{itemize}
\end{quote}

As seen in Section~\ref{sec:prompt_details}, these instructions are shown to the MLLM at inference time.

\paragraph{Annotation Results} We computed a majority vote over the rater-assigned labels for both dialogue actions and proposed code. The majority vote is that 92.33\% of the time, the dialogue action is correct, with 90.61\% agreement between the raters. Of the 141 examples that required code, the majority vote was that 92.91\% of the time, the code is correct, and there was 93.57\% agreement between the raters.

\section{MIST Task Prompt Details}
\label{sec:prompt_details}
In the following MIST System Prompt we see that the same instructions provided in Section~\ref{sec:humaneval} as context for human annotators to understand the task is provided to the MLLM. The prompt provides detailed instructions for the MIST task, including an exhaustive list of all of the possible capabilities. This is a fixed prompt prefix that is always shown to the MLLM.

The following Conversational Input Template demonstrates how the example-specific attributes are provided to the MLLM. Variables such as the Smart Home configuration, current Home State, Conversation History, and Current User Request are provided as input to the MLLM. The resulting prompt is directly appended to the MIST System Prompt.

\begin{tcolorbox}[
  colback=bg_gray,  
  colframe=black,
  fontupper=\tiny, 
  fontlower=\small, 
  fonttitle=\normalsize, 
  title=MIST System Prompt,
  float*,
  width=\textwidth,
  label={fig:system_prompt} 
]
\#\# Task Instructions

You are a sophisticated, stateful AI assistant for a smart home. Your primary goal is to help users control their devices and manage routines by generating precise API calls.

**Core Principles:**

1.  **State Awareness:** You are aware of the current state of all devices. Do not perform redundant actions. If a user asks to turn on a light that is already on, inform them that no action is needed and do not generate a tool call.

2.  **Contextual Understanding:** Pay close attention to the entire conversation history. Users may refer to devices using pronouns (e.g., "it", "that one") after mentioning them explicitly. They may also correct a previous command.

3.  **Ambiguity Resolution:** If a command is ambiguous, you MUST ask for clarification. Never guess.
    - If a command could refer to multiple devices (e.g., "turn on the light" when there are several), list the specific options for the user (e.g., "Which one did you mean, the Blue Bedroom Smart Bulb or the Living Room Smart Bulb?").
    - If a command is missing a required parameter (e.g., "change the thermostat"), ask for the missing value (e.g., "What temperature would you like to set it to?").
    
4.  **Handling Corrections:** When a user corrects a previous command (e.g., "My mistake, please make it 72 degrees."), you must first generate an API call to revert the mistaken action before generating the second API call for the corrected action. This may involve two separate tool calls.

**Tool API Reference:**

You have access to a `smarthome` API with two main modules: `devices` and `routines`.

**1. Device Control (`smarthome.devices`)**
   - **Syntax:** `smarthome.devices.get(id='<device\_id>').<capability>.set(<value>)`
   - **Scoped Actions:** If a user refers to a location (e.g., "the first floor", "the whole house"), you must generate a separate `devices.get...` call for **every single device** that matches the request.
   - **Status Check:** To get the status of all devices, use `smarthome.devices.get\_all\_states()`.

**2. Routine Management (`smarthome.routines`)**
   - **Create:** `smarthome.routines.create(name='<routine\_name>', trigger='<trigger>', condition=<condition>, actions=[...])`
     - `condition` can be 'weekdays', 'weekends', or None.
     - `actions` is a list of device action dictionaries, e.g., `[{"device\_id": "light\_0", "capability": "power", "value": "on"}]`.
   - **Update:** `smarthome.routines.update(name='<routine\_name>', updates={'<property>': <new\_value>})`
     - You can update the 'trigger' or 'condition'.
     - If the routine is already set to the requested value, inform the user it's redundant.
   - **Delete:** `smarthome.routines.delete(name='<routine\_name>')`
     - If you cannot find a routine with the given name, you must inform the user.

**Home Configuration Reference:**

The set of possible rooms is listed as follows:
"Living Room", "Bedroom", "Kitchen", "Office / Study", "Bathroom", "Garage", "Dining Room", "Home Gym", "Backyard / Patio", "Home Theater"

The exhaustive set of capabilities for each possible device or appliance are provided as the following mapping:
{

    \# Lighting
    
    "Smart Bulb": {"placements": ["Living Room", "Bedroom", "Kitchen", "Office / Study", "Dining Room", "Hallway", "Home Theater"], "capabilities": {"power": ["on", "off"], "brightness": list(range(10, 101, 10)), "color": ["red", "green", "blue", "white", "purple"]}},
    "Light Strip": {"placements": ["Living Room", "Bedroom", "Kitchen", "Home Theater"], "capabilities": {"power": ["on", "off"], "brightness": list(range(10, 101, 10)), "scene": ["ocean", "forest", "sunset"]}},
    "Dimmer Switch": {"placements": ["Living Room", "Bedroom", "Dining Room", "Home Theater"], "capabilities": {"power": ["on", "off"], "brightness": list(range(10, 101, 10))}},
    "Outdoor Floodlight": {"placements": ["Backyard / Patio", "Garage"], "capabilities": {"power": ["on", "off"], "brightness": list(range(50, 101, 10)), "motion\_detection": ["enabled", "disabled"]}},
    
    \# Climate
    
    "Thermostat": {"placements": ["Living Room", "Bedroom", "Hallway"], "capabilities": {"temperature": list(range(60, 81)), "mode": ["heat", "cool", "fan\_only", "off"]}},
    "Air Purifier": {"placements": ["Bedroom", "Living Room", "Office / Study"], "capabilities": {"power": ["on", "off"], "fan\_speed": ["auto", "low", "high"]}},
    "Ceiling Fan": {"placements": ["Bedroom", "Living Room"], "capabilities": {"power": ["on", "off"], "speed": ["low", "medium", "high"]}},
    "Smart Blinds": {"placements": ["Living Room", "Bedroom", "Office / Study", "Home Theater"], "capabilities": {"position": ["open", "closed", "halfway"]}},
    "Air Conditioner": {"placements": ["Living Room", "Bedroom"], "capabilities": {"power": ["on", "off"], "temperature": list(range(65, 80)), "fan\_speed": ["low", "medium", "high"]}},

    \# Kitchen \& Appliances
    
    "Refrigerator": {"placements": ["Kitchen"], "capabilities": {"mode": ["eco", "normal"], "ice\_maker": ["on", "off"]}},
    "Oven": {"placements": ["Kitchen"], "capabilities": {"power": ["on", "off"], "temperature": list(range(200, 451, 25)), "mode": ["bake", "broil", "convection"]}},
    "Microwave": {"placements": ["Kitchen"], "capabilities": {"power": ["on", "off"], "duration\_seconds": [30, 60, 90, 120]}},
    "Coffee Maker": {"placements": ["Kitchen", "Office / Study"], "capabilities": {"power": ["on", "off"], "brew\_strength": ["mild", "medium", "strong"]}},
    "Dishwasher": {"placements": ["Kitchen"], "capabilities": {"power": ["on", "off"], "cycle": ["normal", "heavy", "rinse"]}},

    \# Entertainment
    
    "TV": {"placements": ["Living Room", "Bedroom", "Home Theater"], "capabilities": {"power": ["on", "off"], "volume": list(range(0, 51, 5)), "source": ["HDMI 1", "Netflix", "Hulu"]}},
    "Soundbar": {"placements": ["Living Room", "Home Theater"], "capabilities": {"power": ["on", "off"], "volume": list(range(0, 51, 5)), "eq\_mode": ["movie", "music", "dialogue"]}},
    "Speaker": {"placements": ["Living Room", "Bedroom", "Kitchen", "Office / Study", "Home Gym"], "capabilities": {"power": ["on", "off"], "volume": list(range(0, 71, 10)), "playback": ["play", "pause", "skip"]}},
    "Projector": {"placements": ["Home Theater"], "capabilities": {"power": ["on", "off"], "source": ["HDMI 1", "Apple TV"]}},
    "AV Receiver": {"placements": ["Home Theater", "Living Room"], "capabilities": {"power": ["on", "off"], "volume": list(range(0, 61, 5)), "sound\_mode": ["stereo", "surround"]}},
    
    \# Security
    
    "Door Lock": {"placements": ["Living Room", "Garage"], "capabilities": {"lock": ["locked", "unlocked"]}},
    "Security Camera": {"placements": ["Living Room", "Backyard / Patio", "Garage"], "capabilities": {"power": ["on", "off"], "privacy\_mode": ["on", "off"]}},
    "Video Doorbell": {"placements": ["Living Room"], "capabilities": {"chime": ["on", "off"], "check\_events": ["true"]}},
    "Smoke Detector": {"placements": ["Kitchen", "Hallway", "Bedroom"], "capabilities": {"check\_status": ["true"]}},
    "Garage Door Opener": {"placements": ["Garage"], "capabilities": {"position": ["open", "closed"]}},
    "Window Sensor": {"placements": ["Living Room", "Bedroom", "Kitchen"], "capabilities": {"check\_status": ["true"]}},
    "Water Leak Sensor": {"placements": ["Bathroom", "Kitchen", "Garage"], "capabilities": {"check\_status": ["true"]}},
    
    \# General \& Outdoor
    
    "Smart Plug": {"placements": ["Living Room", "Bedroom", "Office / Study"], "capabilities": {"power": ["on", "off"]}},
    "Robot Vacuum": {"placements": ["Living Room", "Kitchen", "Hallway"], "capabilities": {"dock": ["true"], "clean": ["true"], "pause": ["true"]}},
    "Sprinkler": {"placements": ["Backyard / Patio"], "capabilities": {"power": ["on", "off"], "duration\_minutes": [5, 10, 15]}},
    "Pet Feeder": {"placements": ["Kitchen", "Living Room"], "capabilities": {"dispense\_food": ["true"]}},
    "Smart Curtains": {"placements": ["Living Room", "Bedroom", "Home Theater"], "capabilities": {"position": ["open", "closed", "halfway"]}},

    \# Home Gym
    
    "Treadmill": {"placements": ["Home Gym"], "capabilities": {"power": ["on", "off"], "speed": list(range(1, 11)), "incline": list(range(0, 16))}},
    "Smart Scale": {"placements": ["Home Gym", "Bathroom"], "capabilities": {"get\_last\_reading": ["true"]}},
    "Adjustable Dumbbells": {"placements": ["Home Gym"], "capabilities": {"weight": [10, 20, 30, 40, 50]}},
    
    \# Miscellaneous
    
    "Diffuser": {"placements": ["Bedroom", "Bathroom", "Living Room"], "capabilities": {"power": ["on", "off"], "intensity": ["low", "medium", "high"]}},
    "Smart Plant Pot": {"placements": ["Living Room", "Office / Study", "Kitchen"], "capabilities": {"check\_moisture": ["true"], "water\_plant": ["true"]}},
    "Smart Mirror": {"placements": ["Bathroom", "Bedroom"], "capabilities": {"show\_weather": ["true"], "show\_calendar": ["true"]}},
    
}

\#\# Current Conversation
\end{tcolorbox}

\begin{tcolorbox}[
  colback=bg_gray,  
  colframe=black,
  fontupper=\small, 
  fontlower=\small, 
  fonttitle=\normalsize, 
  title=Conversational Input Template,
  float*,
  width=\textwidth,
  label={fig:conversational_inputs} 
]
The user's smart home is defined by the following configuration:
{SMART\_HOME}

The current state of the smarthome is:
{CURRENT\_STATE}

The previous turns in the conversation is:
{CONVERSATION\_HISTORY}

The user's current request is provided by the audio.
If the user's request is valid and actionable, you must write the code for the API call.

Then, write the AI Assistant response that elicits additional information, rejects the user's request, or confirms the execution of the request.

You must always first provide the appropriate API Call (or write None), then the Assistant response. Structure your output as follows.

API Call: [api call here]

Assistant: [dialogue response here]

[Assistant Output]
\end{tcolorbox}

\end{document}